%% file: VER_arxiv.tex
\documentclass[10pt,twocolumn,letterpaper]{article}

\usepackage{cvpr}              
\usepackage{amsmath} 
\usepackage{amssymb}

\usepackage{times} 
\usepackage{bbm} 
\usepackage{bm}

\usepackage{graphicx} 

\usepackage{color}
\definecolor{mygray}{gray}{.9}
\usepackage{colortbl} 
\usepackage{multirow}
\makeatletter
\newcommand{\thickhline}{%
    \noalign {\ifnum 0=`}\fi \hrule height 1pt
    \futurelet \reserved@a \@xhline
}
\makeatother

\usepackage{caption} 

\usepackage[export]{adjustbox} 

\usepackage{enumitem}

\usepackage{algorithm}
\usepackage{listings} 
\usepackage{array}
\usepackage{makecell}
\input{preamble}

\definecolor{cvprblue}{rgb}{0.21,0.49,0.74}
\usepackage[pagebackref,breaklinks,colorlinks,citecolor=cvprblue]{hyperref}

\def\confName{CVPR}
\def\confYear{2024}

\title{Volumetric Environment Representation for Vision-Language Navigation}

\author{{Rui Liu
\quad Wenguan Wang
\quad Yi Yang} \\
\small{ReLER, CCAI, Zhejiang University} \\
\small \url{https://github.com/DefaultRui/VLN-VER}
}

\begin{document}
\maketitle
\begin{abstract}
Vision-language navigation (VLN) requires an agent to navigate through an 3D environment based on visual observations and natural language instructions. It is clear that the pivotal factor for successful navigation lies in the comprehensive scene understanding. Previous VLN agents employ monocular frameworks to extract 2D features of perspective views directly. Though straightforward, they struggle for capturing 3D geometry and semantics, leading to a partial and incomplete environment representation. To achieve a comprehensive 3D representation with fine-grained details, we introduce a Volumetric Environment Representation (VER), which voxelizes the physical world into structured 3D cells. For each cell, VER aggregates multi-view 2D features into such a unified 3D space via 2D-3D sampling. Through coarse-to-fine feature extraction and multi-task learning for VER, our agent predicts 3D occupancy, 3D room layout, and 3D bounding boxes jointly. Based on online collected VERs, our agent performs volume state estimation and builds episodic memory for predicting the next step. Experimental results show our environment representations from multi-task learning lead to evident performance gains on VLN. Our model achieves state-of-the-art performance across VLN benchmarks (R2R, REVERIE, and R4R).
\end{abstract}

\section{Introduction}
Vision-language$_{\!}$ navigation$_{\!}$ (VLN)$_{\!}$ requires$_{\!}$ an$_{\!}$ agent$_{\!}$ to$_{\!}$ navigate$_{\!}$ in$_{\!}$ a 3D$_{\!}$ environment$_{\!}$ following$_{\!}$ natural$_{\!}$ language$_{\!}$ instructions \cite{AndersonWTB0S0G18,qi2020reverie}.$_{\!}$ As a$_{\!}$ holistic$_{\!}$$_{\!}$ understanding of$_{\!}$ the$_{\!}$$_{\!}$ environment$_{\!}$ plays a pivotal role in decision-making within VLN, environment$_{\!}$ representation$_{\!}$ learning$_{\!}$ serves$_{\!}$ as$_{\!}$ a$_{\!}$ foundation$_{\!}$ for formulating accurate navigation policies.

\begin{figure}[t]
	\begin{center}
		\includegraphics[width=0.99\linewidth]{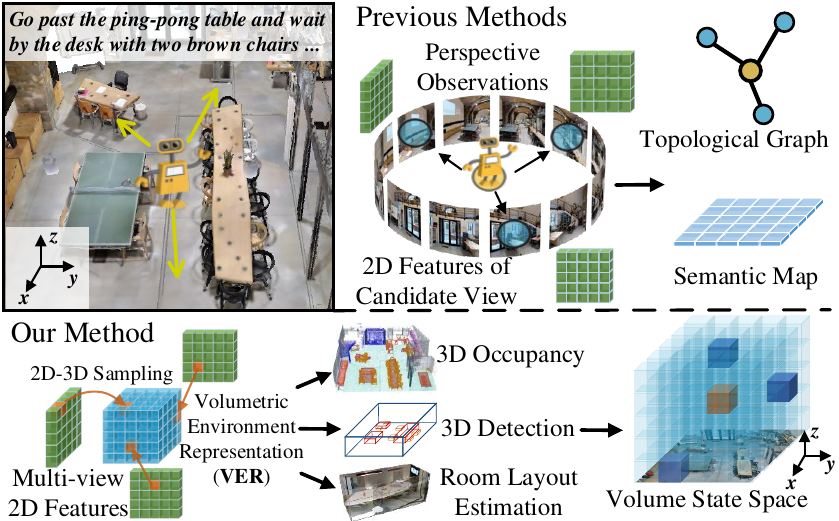}
	\end{center}
	\vspace{-15pt}
	\captionsetup{font=small}
	\caption{\small{The agent observes its surroundings with corresponding perspective features of different candidate views (\protect\includegraphics[scale=0.16,valign=c]{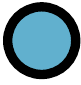}). Previous methods construct the topological graph or semantic map based on these 2D features. Our VER aggregates the multi-view features into structured 3D cells via 2D-3D sampling. VER is a powerful representation for both 3D perception tasks and VLN, providing a volume state space for decision-making.}}
	\label{fig_introduce}
	\vspace{-6pt}
\end{figure}

Early VLN approaches \cite{AndersonWTB0S0G18,fried2018speaker} typically learn the navigation policy through the sequence-to-sequence (Seq2Seq) framework~\cite{sutskever2014sequence}, which directly maps instructions and multi-view perspective observations to actions. They simply embed their immediate observation of local environment into the hidden states of recurrent units. As a result, they lack of explicit environment representations and struggle to access their past states during long-time exploration \cite{wang2021structured,parisotto2018neural}. To address this issue, later VLN agents are equipped with an external memory module \cite{parisotto2018neural}, which stores the environment representations distinctly from navigation states. In this way, they can explicitly model and maintain the environment layouts and contents in a form of topological graph \cite{deng2020evolving,hong2020language,pashevich2021episodic,chen2021topological,chen2022think,an2023etpnav} or semantic map \cite{anderson2019chasing,an2023bevbert,liu2023bird,chen2023omnidirectional,georgakis2022cross,zubair2021sasra,chen2022weakly,hong2023learning,wang2023gridmm,tan2022self} (Fig.~\ref{fig_introduce}). Despite their promising performance with advanced frameworks (\eg, graph neural network~\cite{kipf2016semi} and Transformer~\cite{vaswani2017attention}), their environment representations are still built upon 2D perspective features extracted by monocular frameworks. While straightforward, they compress depth information onto the perspective plane, sacrificing the integral scene structure in the 3D space. Thus, they encounter challenges in capturing 3D geometry and semantics in complex scenes. Such an incomplete environment representation easily leads to sub-optimal navigation decisions.

In this article, we propose a Volumetric Environment Representation (VER) that quantizes the physical world into structured 3D cells (Fig.~\ref{fig_introduce}). These cells, arranged within a predefined volumetric space, maintain both height and depth dimensions. Each cell corresponds to local context of the 3D space. VER aggregates multi-perspective 2D features within these cells through an \textit{environment encoder}~(\S\ref{sec_verencode}). Compared to previous partial representations derived from hidden states and external memory, our VER captures the full geometry and semantics of the physical world. These 3D cells stores the properties of the corresponding space in the scene by predicting 3D occupancy~\cite{tian2023occ3d,song2017semantic}, room layout~\cite{zou2021manhattan}, and 3D object boxes~\cite{li2022bevformer}. However, directly reconstructing the high-quality VER from 2D perspective views is challenging to capture the fine-grained details. As a response, we propose a \textit{coarse-to-fine} VER extraction architecture, which uses learnable up-sampling operations to construct the representations progressively. It is supervised by multi-resolution semantic labels at different scales, utilizing the coarse-to-fine representations as hierarchical inputs. The annotations of the 3D tasks are collected for multi-task learning (\S\ref{sec_benchmark}).

At each navigation step, our agent initially encodes the multi-view observations into VER (\S\ref{sec_stateestimate}). With VER, instructions can be more effectively grounded in the 3D context. This is achieved by establishing cross-modal correlations between linguistic words and 3D cells of VER. Based on the correlations, a \textit{volume state estimation} module is proposed to calculate transition probabilities over the surrounding cells. With the help of this module, our agent performs comprehensive decision-making in volumetric space, and then maps the volume state into local action space. In addition, an \textit{episodic memory} module is established to online collect the information of observed viewpoints and build a topological graph providing global action space (\S\ref{sec_action}). The node embeddings in the graph are from neighbor pillar representations in VER corresponding to the respective viewpoints. To balance the long-range action reasoning and language grounding, our agent combines both the local action probabilities derived from the volume state and the global action probabilities obtained from the episodic memory.

Our agent is evaluated on three VLN benchmarks, \ie, R2R~\cite{AndersonWTB0S0G18}, REVERIE~\cite{qi2020reverie}, and R4R~\cite{jain2019stay} (\S\ref{ex_vln}). It yields solid performance gains (about $\textbf{3}\%$ SR and $\textbf{4}\%$ SPL on R2R \textit{test}, $\textbf{4}\%$ SR and $\textbf{4}\%$ SPL on REVERIE \textit{val unseen}). The ablation study confirms the efficacy of core model designs (\S\ref{ex_ablation}). Additional results show our model achieves promising performance in 3D occupancy prediction, 3D detection, and room layout estimation (\S\ref{ex_perception}).

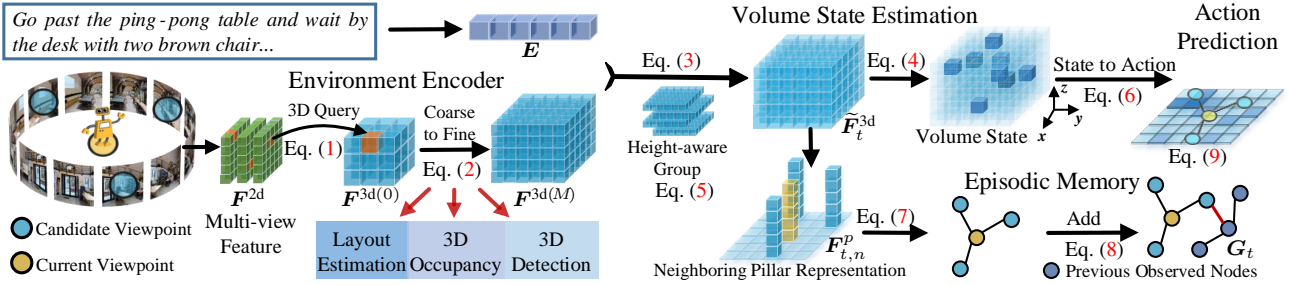
\begin{figure*}[t]
		\vspace{-10pt}
	\begin{center}
		\includegraphics[width=0.98\linewidth]{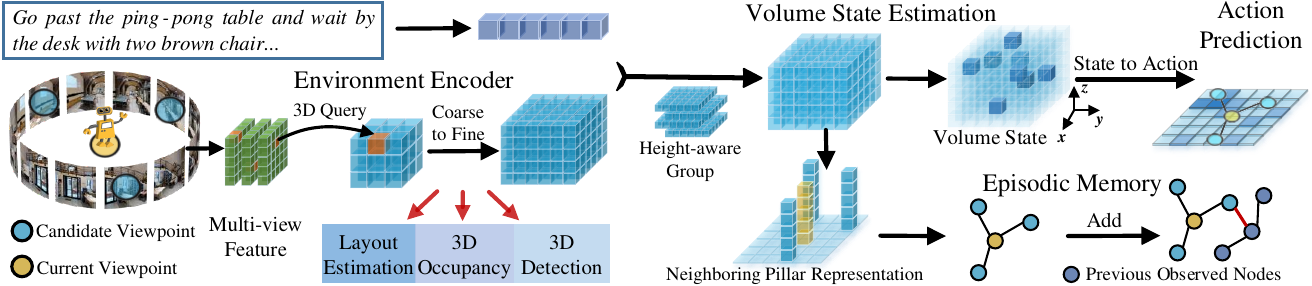}
        \put(-400,28.5){\footnotesize{$\bm{F}^{\text{2d}}$}}
        \put(-379.5,48){\footnotesize{Eq.~(\ref{eq_sample})}}
        \put(-358,28.5){\footnotesize{$\bm{F}^{\text{3d}(0)}$}}
        \put(-293,28.5){\footnotesize{$\bm{F}^{\text{3d}(\!M\!)}$}}
        \put(-327,40){\footnotesize{Eq.~(\ref{eq_upsample})}}
        \put(-290,85){\footnotesize{$\bm{E}$}}
        \put(-244,81){\footnotesize{Eq.~(\ref{voxelatt})}}
        \put(-239,32){\footnotesize{Eq.~(\ref{eq_planeatt})}}
        \put(-169,56){\footnotesize{${\widetilde{\bm{F}}^{\text{3d}}}_t$}}
        \put(-159,81){\footnotesize{Eq.~(\ref{eq_stateestimate})}}
        \put(-77,68){\footnotesize{Eq.~(\ref{eq_localaction})}}
        \put(-175,11){\footnotesize{$\bm{F}^{p}_{t,n}$}}
        \put(-163,22){\footnotesize{Eq.~(\ref{eq_pillar})}}
        \put(-84,9.5){\footnotesize{Eq.~(\ref{eq_globalaction})}}
        \put(-45,45){\footnotesize{Eq.~(\ref{eq_fuseaction})}}
        \put(-23,10){\footnotesize{$\bm{G}_t$}}

	\end{center}
	\vspace{-5pt}
	\captionsetup{font=small}
	\caption{\small{Overview of our model. Given the perspective features of candidate views, a group of 3D queries are used to sample and aggregate them into VER (\S\ref{sec_verencode}). To encode VER, we adopt \textit{coarse-to-fine extraction} and perform multi-task learning on 3D perception. Based on VER, a \textit{volume state estimation} module is proposed to predict state transition (\S\ref{sec_stateestimate}). The \textit{episodic memory} is used to store past observations using neighboring pillar representations for each viewpoint (\S\ref{sec_action}). For decision-making, our agent combines both the local action probabilities from the volume state and the global action probabilities obtained from the episodic memory. See \S\ref{sec_approach} for more details.}}
	\label{fig_framework}
	\vspace{-1pt}
\end{figure*}

\section{Related Work}
\noindent\textbf{Vision-Language Navigation (VLN).} Early VLN agents \cite{AndersonWTB0S0G18,fried2018speaker} are built upon Seq2Seq framework to reserve the observation history in hidden state. Thus they struggle to capture long-range context as the path length increases. Later efforts are devoted to multimodal representation learning, navigation strategy learning, and data generation \cite{gu2022vision}. As a primary step, \textit{multimodal representation learning} helps agents understand the environments and establish relations between the instructions and visual observations. Inspired by the success of vision-language pretraining~\cite{majumdar2020improving,qiao2022hop,tan2019lxmert}, recent approaches~\cite{chen2021history,hong2021vln} use transformer-based architectures~\cite{kenton2019bert,vaswani2017attention} for joint visual and textual representations. Some attempts further exploit the visual information by modeling semantic relation \cite{huo2023geovln} and spatial information \cite{deng2020evolving,hong2020language,chen2021topological,wang2023active,wang2022towards,zhao2022target}. For \textit{navigation strategy learning}, many VLN models~\cite{wang2019reinforced,gao2023adaptive} use imitation and reinforcement learning-based training strategies. Previous solutions \cite{koh2021pathdreamer,wang2023dreamwalker} introduce world models \cite{ha2018recurrent} to perform mental simulations and make mental planning. Furthermore, the scarcity of human instructions and limited diversity of the scene hinder the agent to learn navigation policy and generalize to unseen environments well~\cite{liu2021vision}. Therefore, several VLN \textit{data generation} strategies have been proposed to create new trajectories from existing datasets~\cite{guhur2021airbert,chen2022learning,wang2023scaling}, generate more instructions~\cite{tan2019learning,kamath2023new,wang2023learning,wang2022counterfactual}, or create synthetic environments~\cite{li2022envedit,li2023panogen}. In addition, driven by large models \cite{bubeck2023sparks}, existing VLN agents \cite{zhou2023navgpt,chen20232} demonstrate promising zero-shot performance.

Despite their outstanding contributions, most of them rely on 2D visual cues in perspective observations. These representations are constrained by occlusion and limited geometric information, especially in complex scenes. In this paper, we propose VER, a unified environment representation learned by 3D perception tasks. During navigation, our agent performs volume state estimation on VER, facilitating comprehensive decision-making within the 3D space.

\noindent\textbf{Environment Representation.} Existing VLN models introduce various representations for environment modeling, including topological graphs \cite{wang2021structured,deng2020evolving,hong2020language,chen2021topological,chen2022think,an2023etpnav}, and semantic spatial representations \cite{anderson2019chasing,an2023bevbert,liu2023bird,georgakis2022cross,zubair2021sasra,chen2022weakly,hong2023learning,wang2023gridmm,tan2022self}.
With a broader view, diverse representations have been proposed for robotics and autonomous driving \cite{garg2020semantics,thrun2002probabilistic,li2022delving,wu2021vector,roldao20223d}. In the early stage, 2D occupancy grid maps \cite{elfes2013occupancy} model occupied and free space in the surroundings based on Bayesian estimation for robot navigation. Classic SLAM systems \cite{durrant2006simultaneous} construct a map directly by integrating information from various sensors, including LiDAR and cameras. However, the representations in SLAM still rely on primitives such as 3D point clouds and image patches. Some efforts \cite{cartillier2021semantic,henriques2018mapnet,chaplot2019learning} focus on developing learnable semantic map representations. To enhance spatial reasoning, scene graph representations \cite{chaplot2020neural,wald2020learning} define the topological relations between spatial elements of the environment. In addition, neural scene representations \cite{kwon2023renderable,li20223d} embed image observations into latent codes for object categories, showcasing scalability to large scenes \cite{jiang2020local,mildenhall2021nerf}.
Vision-centric BEV perception, which transforms perspective-view inputs to BEV grid representations, has recently received increasing attention~\cite{philion2020lift,li2022bevformer,li2022delving}. As the BEV representations simplify the vertical geometry, 3D occupancy prediction~\cite{huang2023tri,tong2023scene} is further proposed to infer the 3D geometry from perspective images \cite{xu2020learning,song2017semantic,roldao20223d,wang2023embodiedscan}.

However, existing VLN agents mainly employ abstract relations or compressed spatial maps, lacking the ability to access complete scene information. Providing more world context can be beneficial for the subsequent decision-making and policy learning. Motivated by this insight, we explore a holistic environment representation VER that voxelizes the world into structured 3D cells. VER captures both semantic information and geometric details of the whole scene. Building upon VER, our agent is able to predict the 3D occupancy, room layout, and 3D boxes accurately.

\vspace{10pt}
\section{Approach}\label{sec_approach}
\noindent\textbf{Problem Formulation.} For brevity, we present the technical description in the context of R2R~\cite{AndersonWTB0S0G18}. The navigable area of the environment is organized as an undirected graph, containing a set of nodes (viewpoints) and connectivity edges. In R2R, an embodied agent needs to navigate to a target location in the 3D environment following human instructions with $L$ words (embedded as $\bm{E}\!\!\in\!\!\mathbb{R}^{{D_w}\!\times\! L}$, where $D_w$ is the channel dimension). At time step $t$, the agent looks around and obtains multi-view observations of its surrounding scene from the current location. Each view is represented by a 2D visual feature $\bm{F}^{\text{2d}}_t{_{\!}}\!\in\!\!\mathbb{R}^{{D_i}\!{\times\!{H}\!\times\!{W}}}$, where $H$ and $W$ are the spatial shape of image plane, $D_i$ denotes the channel dimension. The local action space $\mathcal{A}_t\!\!\in\!\!\mathbb{R}^{{N_t}{_{\!}}+\!1}$ is defined by $N_t$ candidate views, which correspond to neighboring navigable nodes $\{v^*_{t,n}\}_{n=1}^{N_t}$, as well as a [STOP] action. Previous agents predict the action probabilities $\bm{p}^{\text{2d}}_t{_{\!}}\!\in{_{\!}}\!\mathbb{R}^{{N_t}{_{\!}}+{_{\!}}1}$ directly based on $\bm{F}^{\text{2d}}_t$ of each candidate view. However, these 2D features with limited geometric information are partial representations of the 3D environment, easily leading to suboptimal decision making.

\noindent\textbf{Overview.} To achieve comprehensive scene understanding,

\noindent we introduce VER, which voxelizes the 3D world into structured 3D cells (Fig.~\ref{fig_framework}). At step $t$, an \textit{environment encoder} is proposed to sample multi-view features ($\bm{F}^{\text{2d}}_t$ of each view) into the volumetric space of VER, forming a unified representation $\bm{F}^{\text{3d}}_t{_{\!}}\!\in{_{\!}}\!\mathbb{R}^{{D_e}\!\times\!{X\!\times\! Y\!\times\! Z}}$ (\S\ref{sec_verencode}). $X$ and $Y$ are the shapes of horizontal plane, $Z$ reserves the height information of 3D space, and $D_e$ represents the channel dimension. The volumetric space aligns with gravity in the world coordinate system based on the Manhattan assumption~\cite{zou2021manhattan}. To encode VER, we devise \textit{coarse-to-fine extraction} with multiple 3D perception tasks supervised by multi-resolution annotations (\S\ref{sec_benchmark}). Based on VER, a \textit{volume state estimation} module is proposed to predict state transition probabilities $\bm{p}^{\text{3d}}_t\!\!\in\!\!\mathbb{R}^{X\!\times\!Y\times\!Z}$ over surrounding 3D cells (\S\ref{sec_stateestimate}). With this module, our agent performs comprehensive decision-making in the 3D space and maps $\bm{p}^{\text{3d}}_t$ to $\bm{p}^{\text{2d}}_t$. To predict the next step, our agent combines both the local action probabilities derived from the volume state and the global action probabilities obtained from the \textit{episodic memory} (\S\ref{sec_action}).

\subsection{Environment Encoder}\label{sec_verencode}

\noindent\textbf{2D-3D Sampling.} At step $t$, the agent observes its surroundings and acquires the multi-view images. We introduce cross-view attention (CVA) to aggregate their features ($\bm{F}^{\text{2d}}$ for each view) into a unified volumetric representation $\bm{F}^{\text{3d}}$ with a group of learnable volume queries $\bm{Q}\in\mathbb{R}^{{D_e}\!\times\!{X\times Y\times Z}}$ ($t$ is omitted for simplicity). Specifically, for the 3D cell positioned at $(x,y,z)$ within the egocentric world, a single query $\bm{Q}(x,y,z)\in\mathbb{R}^{D_e}$ is used to sample each image feature $\bm{F}^{\text{2d}}$ as:
\vspace{-3pt}
\begin{equation}
\small
\begin{aligned}
\bm{F}^{\text{3d}}{(x,y,z)}\!=\!{\text{CVA}}\big(\bm{Q}(x,y,z),{\bm{F}^{2d}{({h'},{w'})}}\big),
\end{aligned}
\label{eq_sample}
\vspace{-2pt}
\end{equation}
where $(h',w')$ denotes the location of corresponding sam-

\noindent pling point on the image plane. Note that we only show the formulation of a single sampling point for conciseness. Since the sampling strategy of vanilla cross-attention is computationally expensive, the deformable attention \cite{zhu2020deformable,li2022bevformer} is introduced and extended in CVA. In this way, $\bm{Q}{(x,y,z)}$ selectively attends to a set of key sampling points around a reference instead of the entire $\bm{F}^{\text{2d}}$.

\begin{figure}[t]
	\begin{center}
		\includegraphics[width=0.98\linewidth]{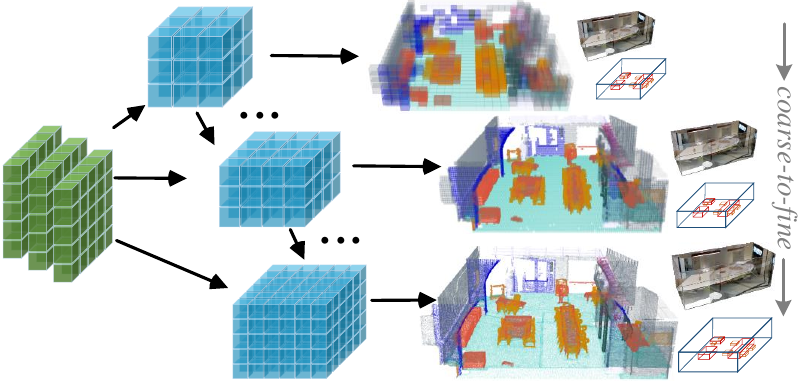}
        \put(-227,20){\footnotesize{$\bm{F}^{\text{2d}}$}}
        \put(-197,49){\footnotesize{Eq.~(\ref{eq_sample})}}
        \put(-197,23){\footnotesize{Eq.~(\ref{eq_sample})}}
        \put(-171,36.8){\footnotesize{Eq.~(\ref{eq_upsample})}}
        \put(-163,80){\footnotesize{$\bm{F}^{\text{3d}(0)}$}}
        \put(-132,3){\footnotesize{$\bm{F}^{\text{3d}(\!\text{M}\!)}$}}

	\end{center}
	\vspace{-7pt}
	\captionsetup{font=small}
	\caption{\small{Our \textit{coarse-to-fine} VER representation extraction (\S\ref{sec_verencode}) adopts cascade up-sampling operations with 3D deconvolutions (Eq.~\ref{eq_upsample}) and 3D queries (Eq.~\ref{eq_sample}). The training process is supervised at different scales by multi-resolution semantic labels.}}
	\label{fig_coarsetofine}
	\vspace{-1pt}
\end{figure}
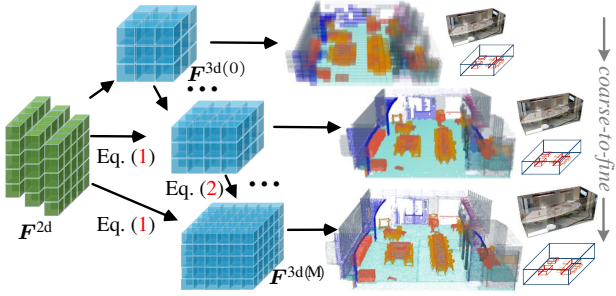

\noindent\textbf{Coarse-to-Fine VER Extraction Architecture.} Directly recovering the fine-grained VER from perspective features easily leads to performance and efficiency degradation~\cite{cherabier2018learning,wei2023surroundocc}. The \textit{coarse-to-fine} extraction is proposed to reconstruct VER progressively. Our approach involves cascade up-sampling operations (Fig.~\ref{fig_coarsetofine}), dividing this extraction into $M$ levels. At each level, 3D deconvolutions are utilized for lifting spatial resolution, and then CVA is used to query the multi-view 2D features (Eq.~\ref{eq_sample}) for refining the detailed geometry. This enables the direct learning of details and avoids the inaccuracy of interpolation \cite{liu2018see,tian2023occ3d,wang2023panoocc} (see Table~\ref{table_coarse2fine}). Between the input coarse feature $\bm{F}^{\text{3d(0)}}\!~\in\!\mathbb{R}^{{D_e}\times \frac{X}{2^{M}}\times \frac{Y}{2^{M}}\times \frac{Z}{2^{M}}}$ from Eq.~(\ref{eq_sample}) and the target fine feature $\bm{F}^{\text{3d}(M)}\in\!\mathbb{R}^{{D_e}\times X\times Y\times Z}$, the intermediate features with varying shapes are calculated as follows:
	\vspace{-3pt}
\begin{equation}
\small
\begin{aligned}
\bm{F}^{\text{3d}(1)}={\uparrow}\bm{F}^{\text{3d}(0)},\cdots,\bm{F}^{\text{3d}(M)}={\uparrow}\bm{F}^{\text{3d}(M-1)},
\end{aligned}
\label{eq_upsample}
	\vspace{-4pt}
\end{equation}
where $\bm{F}^{\text{3d}(1)},\cdots,\bm{F}^{\text{3d}(M-1)}$ denote the intermediate features from different levels, and `$\uparrow$' denotes the up-sampling.

\noindent\textbf{Multi-task Learning.} Our VER offers a unified scene representation for various 3D perception tasks. Existing studies \cite{song2017semantic,roldao20223d} highlight that semantics and geometry are tightly intertwined. We train our environment encoder to extract VER under the supervision of multiple 3D perception tasks (\S\ref{sec_benchmark}). This process utilizes $\{\bm{F}^{\text{3d}(0)},\cdots,\bm{F}^{\text{3d}(M)}\}$ as inputs and is supervised at different scales (Fig.~\ref{fig_coarsetofine}). For 3D occupancy prediction, the decoder is implemented as MLPs with the focal loss~\cite{lin2017focal}. For 3D layout estimation, we adopt a query-based head to yield the manhattan room layouts. A combination of the L1 loss and the IoU loss~\cite{rezatofighi2019generalized} is used as the training objective. For 3D detection, we employ a detection head to predict the 3D boxes \cite{li2022bevformer}. The bipartite matching and the bounding box loss~\cite{li2022bevformer,zhu2020deformable} are employed for detection. A weight vector $[2.0,0.25,0.25]$ is used to balance the optimization of the three tasks, respectively. During navigation, the agent traverses between different viewpoints and encodes VERs through the frozen environment encoder.

\subsection{Volume State Estimation}\label{sec_stateestimate}

VLN task is typically viewed as a state estimation and transition problem \cite{thrun2002probabilistic}. With our VER, the agent state is represented as ``volume state''. As such, the state transition within the locally observed 3D environment, computed by Eq. (\ref{voxelatt})\&(\ref{eq_stateestimate}), is referred as ``volume state estimation''. Different from previous plane-level state models~\cite{an2023bevbert,anderson2019chasing}, VER introduces additional height dimension for 3D state estimation. This enables a more accurate action prediction.

\noindent\textbf{Volume State.} At the beginning of a navigation episode, the agent is located at a start viewpoint $(x_0,y_0,z_0)$. Based on its perception range, a volume state space $\mathcal{X}\!\in\!\mathbb{R}^{X\times Y\times Z}$ is defined corresponding to the 3D physical world with an initial state $s_0\!\!=\!\!(x_0,y_0,z_0)$. At step $t$, the next intermediate state $s_{t+1}\!=\!(x_{t+1},y_{t+1},z_{t+1})$ is determined by the instruction embeddings $\bm{E}$ and VER $\bm{F}^{\text{3d}}_t$ for reaching the goal state $s_T$ ($0\!\!<\!\!t\!\!<\!\!T$). As the entire environment is partially observable, the current state transition ($s_{t}\!\rightarrow\! s_{t+1}$ in $\mathcal{X}$) is regarded as a local consideration for action prediction.

\noindent\textbf{State Estimation.} A \textit{volume state estimation} module is devised to predict the probability distribution $\bm{p}^{\text{3d}}_t\!\in\!\mathbb{R}^{X\times Y\times Z}$ of the intermediate state $s_{t+1}$ conditioned on $\bm{E}$ and $\bm{F}^{\text{3d}}_t$. The environment representation is first reshaped as $\bm{F}^{\text{3d}'}_t\in\mathbb{R}^{{D_e}\times XYZ}$, and then adopt multi-layer transformers (MLT) to model the relations between $\bm{E}$ and $\bm{F}^{\text{3d}'}_t$ as follows:
\vspace{-4pt}
\begin{equation}\small
\begin{aligned}
\widetilde{\bm{F}}^{\text{3d}}_t=\text{MLT}\big([\bm{E};\bm{F}^{\text{3d}'}_t]\big)\in\!\mathbb{R}^{{D_e}\times XYZ},
\end{aligned}
\label{voxelatt}
\vspace{-4pt}
\end{equation}
where $\widetilde{\bm{F}}^{\text{3d}}_t$ is the updated representations, and [;] denotes the concatenation operation. MLT consists of stacked self-attention blocks. Then we use MLPs for state estimation:
\vspace{-8pt}
\begin{equation}\small
\begin{aligned}
\bm{p}^{\text{3d}}_t=\text{Softmax}\big(\text{MLP}(\widetilde{\bm{F}}^{\text{3d}}_t)\big)\in\![0,1]^{X\times Y\times Z}.
\end{aligned}
\label{eq_stateestimate}
\end{equation}

\noindent\textbf{Efficient Height-aware Group.} The computational and memory efficiency of Eq.~(\ref{voxelatt}) is compromised due to the resolution of  $\bm{F}^{\text{3d}}_t$ ($XYZ\!\!\!\!\gg\!\!\!\!L$). Considering the similarity and sparsity of information along the height direction~\cite{wu2023heightformer,zhang2018efficient}, we partition $\bm{F}^{\text{3d}}_t$ into several uniform groups $\{\bm{F}^{g}_{t,z}\in\mathbb{R}^{{D_e}\times XY}\}^{Z}_{z=1}$ along this axis. Then we apply MLT to each group, and Eq.~(\ref{voxelatt}) is reformulated as:
\begin{equation}
\small
\begin{aligned}
\widetilde{\bm{F}}^{g}_{t,i}&=\text{MLT}\big([\bm{E};\bm{F}^{g}_{t,i}]\big)\in\!\mathbb{R}^{{D_e}\times XY},\\
\widetilde{\bm{F}}^{\text{3d}}_t&=\{\widetilde{\bm{F}}^{g}_{t,z}\}^{Z}_{z=1}\in\!\mathbb{R}^{{D_e}\times XYZ},
\end{aligned}
\label{eq_planeatt}
\end{equation}
where the weights of MLT are shared among different groups. The updated features from different groups $\{\widetilde{\bm{F}}^{g}_{t,z}\}^{Z}_{z=1}$ are aggregated along the height to leverage complementary information. For ease of notation, the symbol $\widetilde{\bm{F}}^{\text{3d}}_t$ is slightly reused for the gathered 3D representations. Then, $\bm{p}^{\text{3d}}_t$ is calculated by Eq.~(\ref{eq_stateestimate}).

\subsection{Action Prediction}\label{sec_action}
For action prediction across the entire explored scene, a topological graph $\mathcal{G}_t\!\!=\!\!\{\mathcal{V}_t,\mathcal{E}_t\}$ is constructed and updated online to represent \textit{episodic memory} during navigation. Specifically, $\mathcal{V}_t$ denotes the observed viewpoints, \ie, all visited viewpoints and their candidate viewpoints. These viewpoints are encoded by compressing previous VERs. The edge $\mathcal{E}_t$ denotes the navigable connections between these viewpoints. To predict the next step, our agent combines both the volume state estimation and episodic navigation memory for decision-making.

\noindent\textbf{Mapping Volume State to Action.} As our agent navigates on the horizontal plane to reach the adjacent candidate viewpoints $\{v^*_n\}_{n=1}^{N_t}$, we map the volume state space into 2D space to align with this movement pattern. Specifically, we average $\bm{p}^{\text{3d}}_t$ along the height ($z$-axis) axis as $\bm{p}^{h}_t\!\!\in\!\!\mathbb{R}^{X\times Y}$. Then we sum probability values in the neighborhood of $\{v^*_{t,n}\}_{n=0}^{N_t}$ ($v^*_0$ for the current viewpoint, \ie, [STOP]), and$_{\!}$ normalize$_{\!}$ them$_{\!}$ as$_{\!}$ the$_{\!}$ local$_{\!}$ action$_{\!}$ probabilities:
\vspace{-2pt}
\begin{equation}\small
\begin{aligned}
\bm{p}^{\text{2d}}_t=\!\big\{\sum\!{\bm{p}^{h}_t(x_n,y_n)}\big\}_{n=0}^{N_t}\in\![0,1]^{N_t+1}, (x_n,y_n)\in\Omega_{n},
\end{aligned}\label{eq_localaction}
\vspace{-2pt}
\end{equation}
where $\bm{p}^{h}_t(x_n,y_n)$ is the value at the coordinate $(x_n,y_n)$, and $\Omega_{n}$ is the neighborhood of $v^*_n$ in the horizontal plane. In the training stage, a heatmap \cite{zhou2019objects} with a Gaussian kernel is used to supervise this action prediction (\S\ref{sec_implementation}).

\noindent\textbf{Global Action Prediction.} The \textit{episodic memory} module is used to store past environment representations and allows easy access to them. To memory the environment representations efficiently, we use the neighboring pillar \cite{wang2020pillar} representations $\bm{F}^{p}_{t,n}\!\in\!\mathbb{R}^{{D_e}\times{|\Omega_{n}|}\times Z}$, corresponding to the current observed viewpoints $\{v^*_n\}_{n=0}^{N_t}$ at step $t$:
\vspace{-5pt}
\begin{equation}\small
\begin{aligned}
\bm{F}^{p}_{t,n}=\{\widetilde{\bm{F}}^{\text{3d}}_{t}(x_n,y_n,z_n)\}_{(x_n,y_n)\in\Omega_{n},{z_n}\in[1,Z]}
\end{aligned}
\label{eq_pillar}
\end{equation}
where $\widetilde{\bm{F}}^{\text{3d}}_{t}(x_n,y_n,z_n)$ is the representation at position $(x_n,y_n,z_n)$ of $\widetilde{\bm{F}}^{\text{3d}}_{t}$ (Eq.~\ref{eq_planeatt}). $\{\bar{\bm{F}}^{p}_{t,n}\!\!\in\!\!\mathbb{R}^{D_e}\}_{n=0}^{N_t}$ is obtained by average pooling as the corresponding node embeddings, which are then incorporated into $\mathcal{G}_{t}$. For previously observed nodes, we compute the average of their features.

The episodic memory $\mathcal{G}_t$, which includes the observed viewpoints, offers a global action space $\mathcal{A}^{*}_t\!\in\!\mathbb{R}^{|\mathcal{V}_t|}$. This enables our agent to change its current navigation state easily by `jumping' directly to another viewpoint, which may be even observed several steps ago. The global action probabilities on $\mathcal{G}_t$ are calculated as:
\vspace{-2pt}
\begin{equation}
\small
\begin{aligned}
\hat{\bm{G}}_t&=\text{MLT}\big([\bm{E};\bm{G}_t]\big)\in\!\mathbb{R}^{{D_e}\times |\mathcal{V}_t|},\\
\bm{p}^{g}_t&=\text{Softmax}(\text{MLP}(\hat{\bm{G}}_t))\in\![0,1]^{|\mathcal{V}_t|},\\
\end{aligned}
\vspace{-2pt}
\label{eq_globalaction}
\end{equation}
where $\bm{G}_t$ denotes the node embeddings of $\mathcal{G}_t$. The ultimate action probabilities are given as:
\vspace{-2pt}
\begin{equation}
\small
\begin{aligned}
\hat{\bm{p}}^{\text{2d}}_t&=[\bm{p}^{\text{2d}\rightarrow{g}}_t;\bm{p}^{\text{2d}}_t]\in\![0,1]^{|\mathcal{V}_t|},\\
\hat{\bm{p}}^{g}_t&={W_g}\bm{p}^{g}_t+(1-W_g)\hat{\bm{p}}^{\text{2d}}_t\in\![0,1]^{|\mathcal{V}_t|},
\end{aligned}
\label{eq_fuseaction}
\vspace{-2pt}
\end{equation}
where $\bm{p}^{\text{2d}\rightarrow{g}}_t\!\!\in\!\!\mathbb{R}^{|\mathcal{V}_t|-(N_t+1)}$ denotes the probabilities of global backtracking and we use the same value as the local [STOP] probability in Eq.~(\ref{eq_localaction}); $W_g$ is a learnable weight.

\noindent\textbf{State Transition and Memory Update.} After executing the action in $\mathcal{A}^{*}_t$, our agent reaches the next viewpoint $v^*_{t+1,0}$, and will iteratively: (1) encode its current observation as $\bm{F}^{\text{3d}}_{t+1}$ through Eq.~(\ref{eq_sample}); (2) update its volume state as $s_{t+1}\!=\!(x_{t+1},y_{t+1},z_{t+1})$; (3) add the node embeddings of~$\{v^*_{t+1,n}\}_{n=0}^{N_{t+1}}$ into the episodic memory $\mathcal{G}_{t+1}$; and (4) predict the next step with the updated episodic memory (\ie, $\mathcal{G}_{t+1}$) and volume state (\ie, $s_{t+1}$). Our agent repeats the above process until it chooses the [STOP] action or reaches the maximum step limit.

\subsection{Annotation Generation}\label{sec_benchmark}
A multi-task learning framework is proposed to extract and encode the VERs (\S\ref{sec_verencode}). To achieve this, we generate annotations on Matterport3D dataset~\cite{chang2017matterport3d} for 3D occupancy prediction, object detection, and room layout estimation. We design a \textit{room-object-voxel} pipeline to automatically generate these annotations. This pipeline leverages existing LiDAR point labels without additional human annotations (more details in Appendix). We utilize the egocentric observations with multi-view images as input.

\noindent\textbf{Room Layout.} The room layout in our context specifies the positions, orientations, and heights of the walls, relative to the camera center. It aims to reconstruct cuboid room shapes within the Manhattan world~\cite{coughlan1999manhattan}. Given that the horizontal plane is aligned on the $x\!-\!z$ axis, we parameterize the layout with center coordinate, width, length, height, and rotation. In contrast to directly operating on a single panoramic image \cite{zou2021manhattan,zou2018layoutnet}, we use the embodied observations with multi-view images as input.

\noindent\textbf{Object Detection.} Based on the room layout, we collect the surrounding objects if the agent locates in a room. In a non-closed environment, we collect information about nearby objects based on their distance from the agent. For each object, the eigenvectors of its vertices are used to define an oriented bounding box that tightly encloses the object \cite{hua2016scenenn}. Considering some objects may disappear from view due to occlusion but still exist in the environment (referred to as permanence \cite{xu2020learning}), we also include them in the analysis.

\noindent\textbf{Point Accumulation for Occupancy.} To generate voxel labels for occupancy \cite{wei2023surroundocc,tian2023occ3d}, we accumulate the sparse LiDAR points and utilize 3D boxes. Given dense background and object points, we first voxelize the 3D space and label each voxel based on the majority vote of labelled points in that voxel. Due to the limited number of LiDAR points, we leverage the Nearest Neighbors algorithm to generate dense labels for remaining voxels by searching the nearest semantic label. Moreover, the agent infers the complete 3D occupancy of each object (amodal perception), including regions that are not directly observed. This attribute enables the agent to predict the entire object instead of only visible surfaces \cite{xu2020learning}.

\noindent\textbf{Statistics.} For high-resolution labels, we define $120\times 120\times 35$ voxel grids in world coordinates, where the scene voxel size equals to $0.1$ m (see more details about multi-resolution labels in Appendix). We annotate over $50$ billion voxels and $16$ classes, including $11$ foreground objects and $5$ background stuffs. It comprises about $100k$ annotated bounding boxes and $1,\!500$ room layouts within the scenes. These annotations follow the same \textit{train}/\textit{val}/\textit{test} splits as R2R \cite{AndersonWTB0S0G18}. There are $61$ scenes for \textit{train/val seen}, $11$ scenes for \textit{val unseen}, and $18$ scenes for \textit{test}.

\subsection{Implementation Details}\label{sec_implementation}
Initially, the environment encoder (\S\ref{sec_verencode}) is introduced for VER through \textit{coarse-to-fine} extraction. Then multi-task learning is performed across multiple 3D perception tasks, including 3D occupancy prediction, 3D layout estimation, and 3D detection. During navigation, our agent is equipped with the frozen environment encoder and predicts the next step. Following recent VLN practice~\cite{hong2021vln,chen2021history,chen2022think}, both offline pretraining and finetuning are adopted. In this section, we will
mainly introduce the details of architecture and training (see more details in Appendix).

\noindent\textbf{Environment Representation Learning.} For the multi-view images, we adopt ViT-B/16~\cite{dosovitskiy2020image} pretrained on ImageNet to extract features. The number of 3D volume queries is ${15}\times{15}\times{4}$. For each query, it is projected to sample 2D views according to intrinsic and extrinsic parameters of the camera. We set the perception range as $[-6 m,6 m]$ for $x-y$ axis and $[-1.5 m, 2 m]$ for the height ($z$ axis). We adopt six layers of CVA (Eq.~\ref{eq_sample}) for 2D-3D sampling, and then use $M\!=\!3$ cascade deconvolutions for up-sampling (Eq.~\ref{eq_upsample}). The feature dimension is $768$ (\ie, $D_i\!\!=\!\!D_w\!\!=\!\!D_e\!\!=\!\!768$).

\noindent\textbf{Navigation Network.} MLT with 4 layers is initialized from \cite{tan2019lxmert} for state estimation (Eq.~\ref{eq_stateestimate}) and global action prediction (Eq.~\ref{eq_globalaction}), respectively. The range of neighborhood for each candidate is set as $|\Omega_{n}|=9$. The standard deviation of the Gaussian kernel for the heat map is set as $3.0$. Based on this heat map, the focal loss~\cite{lin2017focal} is used to supervise the local action prediction (Eq.~\ref{eq_localaction}). We also use a cross-entropy loss for the global action prediction (Eq.~\ref{eq_fuseaction}).

\noindent\textbf{Pretraining.} For R2R~\cite{AndersonWTB0S0G18} and R4R~\cite{jain2019stay}, Masked Language Modeling~\cite{kenton2019bert,chen2021history} and Single-step Action Prediction~\cite{chen2021history,hong2021vln} are adopted as auxiliary tasks on offline-sampled instruction-route pairs~\cite{hao2020towards}. For REVERIE~\cite{qi2020reverie}, an additional
Object Grounding (OG)~\cite{chen2022think,lin2021scene} is used for object reasoning. During pretraining, we train the model
with a batch size of $64$ for $100k$ iterations, using Adam~\cite{KingmaB14adam} optimizer with 1e-4 learning rate. Only one task is adopted at each mini-batch with the same sampling ratio.

\noindent\textbf{Finetuning.} Following the standard protocol~\cite{chen2022think,an2023bevbert,wang2023gridmm}, we finetune the navigation network using Dagger~\cite{RossGB11} techniques. In addition, the OG loss~\cite{lin2021scene,chen2022think,an2023bevbert} is employed on REVERIE. In this stage, we set the learning rate to 1e-5 and batch size to $8$ with $20k$ iterations.

\noindent\textbf{Inference.} During the testing phase, the agent receives the multi-view images and encodes them as VERs through the frozen environment encoder~(\S\ref{sec_verencode}). Based on VERs, the agent performs volume state estimation~(\S\ref{sec_stateestimate}) and models episodic memory~(\S\ref{sec_action}). By combining both of them, the agent predicts the next step accurately until stops.

\noindent\textbf{Reproducibility.} Our model is implemented in PyTorch and trained on eight RTX 4090 GPUs with a $24$GB memory per-card. Testing is conducted on the same machine.

\vspace*{3pt}
\section{Experiment}

\begin{table}[t]
\centering
        \resizebox{0.49\textwidth}{!}{
		\setlength\tabcolsep{3pt}
		\renewcommand\arraystretch{1.0}
\begin{tabular}{c||cccc|cccc}
\hline \thickhline
\rowcolor{mygray}
~ &  \multicolumn{8}{c}{R2R} \\
\cline{2-9}
\rowcolor{mygray}
~ & \multicolumn{4}{c|}{\textit{val} \textit{unseen}} & \multicolumn{4}{c}{\textit{test} \textit{unseen}} \\
\cline{2-9}
\rowcolor{mygray}
\multirow{-3}{*}{Models} &TL$\downarrow$ &NE$\downarrow$ &SR$\uparrow$ &SPL$\uparrow$ &TL$\downarrow$ &NE$\downarrow$ &SR$\uparrow$ &SPL$\uparrow$\\
\hline
\hline
Seq2Seq~\cite{AndersonWTB0S0G18}  &8.39    &7.81 &22 &$-$    &8.13  &7.85 &20 &18 \\
SF~\cite{fried2018speaker}        &$-$     &6.62 &35 &$-$    &14.82 &6.62 &35 &28 \\
EnvDrop~\cite{tan2019learning}            &10.70   &5.22 &52 &48     &11.66 &5.23 &51 &47 \\
AuxRN~\cite{zhu2020vision}        &$-$     &5.28 &55 &50     &$-$   &5.15 &55 &51 \\
Active~\cite{wang2020active}   &20.60   &4.36 &58 &40     &21.60 &4.33 &60 &41\\
RecBERT~\cite{hong2021vln}         &12.01   &3.93 &63 &57     &12.35 &4.09 &63 &57 \\
HAMT~\cite{chen2021history}             &11.46   &2.29 &66 &61     &12.27 &3.93 &65 &60 \\
SOAT~\cite{moudgil2021soat}       &12.15   &4.28 &59 &53     &12.26 &4.49 &58 &53 \\
SSM~\cite{wang2021structured}     &20.7    &4.32 &62 &45     &20.4  &4.57 &61 &46 \\
CCC~\cite{wang2022counterfactual} &$-$     &5.20 &50 &46     &$-$   &5.30 &51 &48 \\
HOP~\cite{qiao2022hop}            &12.27   &3.80 &64 &57     &12.68 &3.83 &64 &59 \\
DUET~\cite{chen2022think}            &13.94   &3.31 &72 &60     &14.73 &3.65 &69 &59 \\
LANA~\cite{wang2023lana}          &12.0 &$-$ &68 &62      &12.6  &$-$ &65 &60\\
TD-STP~\cite{zhao2022target}    &$-$  &3.22 &70 &63      &$-$ &3.73 &67 &61\\
BSG~\cite{liu2023bird}            &14.90   &2.89 &74 &62    &14.86 &3.19 &73 &62 \\
BEVBert~\cite{an2023bevbert}      &14.55   &2.81 &75 &64     &$-$   &3.13 &73 &62 \\

\hline
\textbf{Ours}                     &14.83   &2.80 &\textbf{76} &\textbf{65}     &15.23 &\textbf{2.74} &\textbf{76} &\textbf{66} \\
\hline
\end{tabular}
}
	\vspace*{-5pt}
\captionsetup{font=small}
	\caption{\small{Quantitative results on R2R~\cite{AndersonWTB0S0G18} (more details in \S\ref{ex_vln}).}}
    \label{table_R2R}
\vspace*{-5pt}
\end{table}

\begin{table*}[t]
\centering
	\vspace{-10pt}
        \resizebox{1\textwidth}{!}{
		\setlength\tabcolsep{2.5pt}
		\renewcommand\arraystretch{1.0}
\begin{tabular}{c||cccccc|cccccc|cccccc}
\hline \thickhline
\rowcolor{mygray}
~ &  \multicolumn{18}{c}{REVERIE} \\
\cline{2-19}
\rowcolor{mygray}
~ &  \multicolumn{6}{c|}{\textit{val} \textit{seen}} & \multicolumn{6}{c|}{\textit{val} \textit{unseen}} & \multicolumn{6}{c}{\textit{test} \textit{unseen}} \\
\cline{2-19}
\rowcolor{mygray}
\multirow{-3}{*}{Models} &\small{TL$\downarrow$} &\small{OSR$\uparrow$} &\small{SR$\uparrow$} &\small{SPL$\uparrow$} &\small{RGS$\uparrow$} &\small{RGSPL$\uparrow$} &\small{TL$\downarrow$} &\small{OSR$\uparrow$} &\small{SR$\uparrow$} &\small{SPL$\uparrow$} &\small{RGS$\uparrow$} &\small{RGSPL$\uparrow$} &\small{TL$\downarrow$} &\small{OSR$\uparrow$} &\small{SR$\uparrow$} &\small{SPL$\uparrow$} &\small{RGS$\uparrow$} &\small{RGSPL$\uparrow$}\\
\hline
\hline
RCM~\cite{wang2019reinforced}    &10.70 &29.44 &23.33 &21.82 &16.23 &15.36    &11.98 &14.23 &9.29  &6.97  &4.89  &3.89      &10.60 &11.68 &7.84  &6.67  &3.67  &3.14  \\
FAST-M~\cite{qi2020reverie}  &16.35 &55.17 &50.53 &45.50 &31.97 &29.66    &45.28 &28.20 &14.40 &7.19  &7.84  &4.67      &39.05 &30.63 &19.88 &11.61 &11.28 &6.08  \\
SIA~\cite{lin2021scene}          &13.61 &65.85 &61.91 &57.08 &45.96 &42.65    &41.53 &44.67 &31.53 &16.28 &22.41 &11.56     &48.61 &44.56 &30.80 &14.85 &19.02 &9.20  \\
RecBERT~\cite{hong2021vln}        &13.44 &53.90 &51.79 &47.96 &38.23 &35.61    &16.78 &35.02 &30.67 &24.90 &18.77 &15.27     &15.86 &32.91 &29.61 &23.99 &16.50 &13.51 \\
Airbert~\cite{guhur2021airbert}       &15.16 &48.98 &47.01 &42.34 &32.75 &30.01    &18.71 &34.51 &27.89 &21.88 &18.23 &14.18     &17.91 &34.20 &30.28 &23.61 &16.83 &13.28 \\
HAMT~\cite{chen2021history}            &12.79 &47.65 &43.29 &40.19 &27.20 &25.18    &14.08 &36.84 &32.95 &30.20 &18.92 &17.28     &13.62 &33.41 &30.40 &26.67 &14.88 &13.08 \\
HOP~\cite{qiao2022hop}           &13.80 &54.88 &53.76 &47.19 &38.65 &33.85    &16.46 &36.24 &31.78 &26.11 &18.85 &15.73     &16.38 &33.06 &30.17 &24.34 &17.69 &14.34 \\
DUET~\cite{chen2022think}           &13.86 &73.86 &71.75 &63.94 &57.41 &51.14    &22.11 &51.07 &46.98 &33.73 &32.15 &23.03     &21.30 &56.91 &52.51 &36.06 &31.88 &22.06 \\
TD-STP~\cite{zhao2022target}    &$-$ &$-$ &$-$ &$-$ &$-$ &$-$      &$-$ &39.48 &34.88 &27.32 &21.16 &16.56         &$-$ &40.26 &35.89 &27.51 &19.88 &15.40\\
BEVBert~\cite{an2023bevbert}        &$-$   &76.18   &73.72   &65.32   &57.70   &51.73    &$-$   &56.40 &51.78 &36.37 &34.71 &24.44 &$-$   &57.26 &52.81 &36.41 &32.06 &22.09 \\
GridMM~\cite{wang2023gridmm}        &$-$   &$-$   &$-$   &$-$   &$-$   &$-$    &23.20 &57.48 &51.37 &36.47 &34.57 &24.56 &19.97 &59.55 &53.13 &36.60 &34.87 &23.45 \\
LANA~\cite{wang2023lana}         &15.91 &74.28 &71.94 &62.77 &59.02 &50.34    &23.18 &52.97 &48.31 &33.86 &32.86 &22.77     &18.83 &57.20 &51.72 &36.45 &32.95 &22.85\\
BSG~\cite{liu2023bird}              &15.26 &78.36 &76.18 &66.69 &61.56 &54.02    &24.71 &58.05 &52.12 &35.59 &35.36 &24.24     &22.90 &62.83 &56.45 &38.70 &33.15 &22.34 \\

\hline
\textbf{Ours}                        &16.13 &\textbf{80.49} &75.83 &66.19 &\textbf{61.71} &\textbf{56.20}     &23.03 &\textbf{61.09} &\textbf{55.98} &\textbf{39.66} &33.71 &23.70   &24.74 &62.22 &\textbf{56.82} &\textbf{38.76} &33.88 &23.19 \\
\hline
\end{tabular}
}
	\vspace*{-5pt}
\captionsetup{font=small}
	\caption{\small{Quantitative comparison results on REVERIE~\cite{qi2020reverie}. `$-$': unavailable statistics (see \S\ref{ex_vln} for more details).}}
    \label{table_REVERIE}
\vspace*{-5pt}
\end{table*}

\begin{table}[t]
\centering
        \resizebox{0.49\textwidth}{!}{
		\setlength\tabcolsep{7pt}
		\renewcommand\arraystretch{1.0}
\begin{tabular}{c||ccccc}
\hline \thickhline
\rowcolor{mygray}
~ & \multicolumn{5}{c}{R4R \textit{val} \textit{unseen}}  \\
\cline{2-6}
\rowcolor{mygray}
\multirow{-2}{*}{Models}    &NE$\downarrow$ &SR$\uparrow$ &CLS$\uparrow$ &nDTW$\uparrow$ &SDTW$\uparrow$ \\
\hline
\hline
SF~\cite{AndersonWTB0S0G18}        &8.47    &24   &30   &$-$   &$-$  \\
RCM~\cite{wang2019reinforced}     &$-$     &29   &35   &30    &13   \\
EGP~\cite{deng2020evolving}       &8.00    &30   &44   &37    &18   \\
SSM~\cite{wang2021structured}     &8.27    &32   &53   &39    &19   \\
RelGraph~\cite{hong2020language}  &7.43    &36   &41   &47    &34   \\
RecBERT~\cite{hong2021vln}         &6.67    &44   &51   &45    &30   \\
HAMT~\cite{chen2021history}        &6.09    &45   &58   &50    &32   \\
BSG~\cite{liu2023bird}            &6.12    &47   &59   &53    &34    \\
LANA~\cite{wang2023lana}          &$-$     &43   &60   &52    &32\\
\hline
\textbf{Ours}                     &6.10   &\textbf{47} &\textbf{61} &\textbf{54} &33   \\
\hline
\end{tabular}
}
	\vspace*{-5pt}
\captionsetup{font=small}
	\caption{\small{Quantitative results on R4R~\cite{jain2019stay} (more details in \S\ref{ex_vln}).}}
    \label{table_R4R}
\vspace*{-5pt}
\end{table}

\vspace*{-1pt}
\subsection{Performance on VLN}\label{ex_vln}
\vspace*{-1pt}
\noindent\textbf{Datasets.}~The experiments are conducted on three datasets. R2R~\cite{AndersonWTB0S0G18} contains $7,\!189$ trajectories sampled from $90$ real-world indoor scenes. It consists of $22k$ human-annotated navigational instructions. The dataset is split into \textit{train}, \textit{val} \textit{seen}, \textit{val} \textit{unseen}, and \textit{test} \textit{unseen} sets, which mainly focus on the generalization capability in unseen environments. REVERIE~\cite{qi2020reverie} contains high-level instructions which describe target locations and objects, with a focus on grounding remote target objects. R4R~\cite{jain2019stay} is an extended version of R2R with longer trajectories.

\noindent\textbf{Evaluation Metrics.}~For R2R, Success Rate (SR), Trajec- tory Length (TL), Oracle Success Rate (OSR), Success rate weighted by Path Length (SPL), and Navigation Error (NE) are used. For REVERIE, Remote Grounding Success rate (RGS), and Remote Grounding Success weighted by Path Length (RGSPL) are also employed for object grounding. For R4R, Coverage weighted by Length Score (CLS), normalized Dynamic Time Warping (nDTW), and Success rate weighted nDTW (SDTW) are used.

\noindent\textbf{Performance on R2R.}~Table~\ref{table_R2R} compares our
model with the state-of-the-art models on R2R. As we find that our model yields SR of $\textbf{76}$\% and SPL of $\textbf{66}\%$ on \textit{text unseen}, which leads to promising gains of $\textbf{3}\%$ and $\textbf{4}\%$ over BEVBert~\cite{an2023bevbert}, respectively. It verifies that using VER to represent the environment leads to better decision-making.

\noindent\textbf{Performance on REVERIE.}~Table~\ref{table_REVERIE} presents the comparison results on REVERIE. Compared with the recent state-of-the-art VLN agent~\cite{liu2023bird}, our agent improves SR and SPL by $\textbf{3.86}\%$ and $\textbf{4.07}\%$ on the \textit{val unseen} split. This highlights the effectiveness of our architecture design. 

\noindent\textbf{Performance on R4R.}~Table~\ref{table_R4R} shows results on R4R. Our approach outperforms others in most metrics with a promising gain on nDTW (\ie, $\textbf{1}\%$). This suggests the \textit{episodic memory} module is able to retrieve the long-time context.

\begin{figure*}[t]
	\begin{center}
	\vspace{-10pt}
		\includegraphics[width=0.99\linewidth]{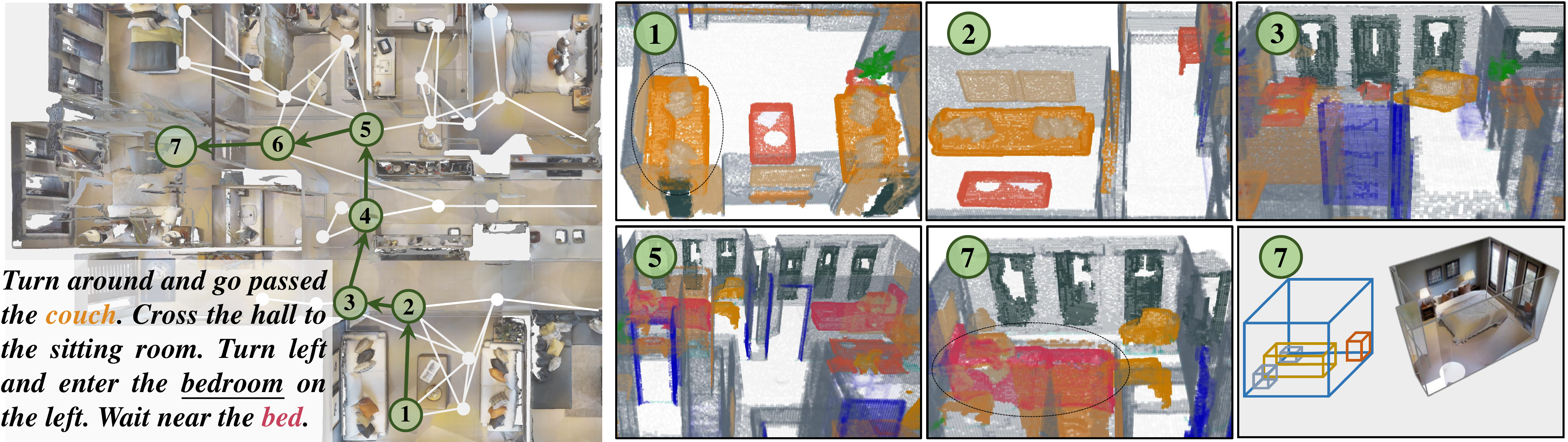}
        \put(-95,6){\footnotesize{3D Boxes}}
        \put(-49,6){\footnotesize{Room Layout}}
	\end{center}
	\vspace{-10pt}
	\captionsetup{font=small}
	\caption{\small{A representative visual result on \textit{val unseen} of R2R~\cite{AndersonWTB0S0G18}. We first visualize the 3D occupancy prediction at the key steps. In addition, we provide the prediction of 3D boxes and 3D room layout at step \protect\includegraphics[scale=0.16,valign=c]{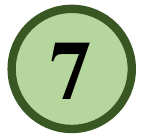}. We find that VER can capture the geometric details of `couch' and the structure of `bedroom'. With VER, our agent easily finds the `bed' and succeeds. See \S\ref{ex_vln} for more details.}}
	\label{fig_result}
	\vspace{-1pt}
\end{figure*}

\noindent\textbf{Visual Results.}~Fig.~\ref{fig_result} depicts one exemplar navigation episode from \textit{val unseen} set of R2R. In this complex environment, there are many rooms with different objects and 3D layout. From the visualization of 3D occupancy prediction at the key steps, we find the geometric details and semantics can be captured well by VER. The room layout estimation can help the agent to understand ``enter the bedroom''. Finally, our agent finds the ``bed'' and accomplishes the instruction successfully.

\vspace{5pt}
\subsection{Diagnostic Experiment}\label{ex_ablation}
To thoroughly test the efficacy of crucial components of our model, we conduct a series of diagnostic studies on \textit{val unseen} split of REVERIE and R2R.

\noindent\textbf{Overall Design.} We first investigate the effectiveness of our core design. The results in Table~\ref{table_overalldesign} indicate that adding \textit{Volume State} leads to a substantial performance gain (\ie, $\textbf{3.67}\%$ on SR). After using \textit{Episodic Memory}, a higher score (\ie, $31.36\%\rightarrow\textbf{33.71}\%$ on RGS) is achieved.

\begin{table}[t]
    \begin{center}
            \resizebox{0.49\textwidth}{!}{
            \setlength\tabcolsep{6.5pt}
            \renewcommand\arraystretch{1.05}
    \begin{tabular}{c|ccc|cc}
    \hline \thickhline
    \rowcolor{mygray}
    & \multicolumn{3}{c|}{REVERIE} & \multicolumn{2}{c}{R2R} \\
    \cline{2-6}
    \rowcolor{mygray}
    \multicolumn{1}{c|}{\multirow{-2}{*}{Models}}         &{SR}$\uparrow$ &SPL$\uparrow$ &{RGS}$\uparrow$ &{SR}$\uparrow$ &SPL$\uparrow$\\
    \hline
    \hline
    ${w/o}$ Volume State                      &52.31 &34.91 &32.75 &72.71 &61.13  \\
    ${w/o}$ Episodic Memory                   &49.33 &33.71 &31.36 &68.21 &61.70  \\
    Full model                                &\textbf{55.98} &\textbf{39.66}  &\textbf{33.71}  &\textbf{75.80} &\textbf{65.37}  \\ \hline
    \end{tabular}
    }
    \end{center}
        \vspace*{-10pt}
    \captionsetup{font=small}
	\caption{\small{Ablation study of overall design on \textit{val} \textit{unseen} of REVERIE~\cite{qi2020reverie} and R2R~\cite{AndersonWTB0S0G18} (see \S\ref{ex_ablation} for more details).}}
    \label{table_overalldesign}
    \vspace*{-8pt}
\end{table}

\noindent\textbf{Neighborhood Range of Viewpoints.} We use the neighborhood of each viewpoint in the state space for local action prediction (Eq.~\ref{eq_localaction}). Moreover, corresponding pillar representations of the neighborhood are also used for node embeddings of the episodic memory (Eq.~\ref{eq_pillar}). From Table~\ref{table_neighborhood}, the limited range of neighborhood is insufficient to represent the candidate viewpoint for navigation (\eg, $\textbf{75.80}\%\rightarrow73.75\%$ of SR on R2R). However, too large neighborhood range will contain irrelevant information, leading to inferior performance (\eg, $\textbf{55.98}\%\!\!\rightarrow\!\!53.49\%$ of SR on REVERIE).

\begin{table}[t]
    \begin{center}
            \resizebox{0.48\textwidth}{!}{
            \setlength\tabcolsep{11pt}
            \renewcommand\arraystretch{1.05}
    \begin{tabular}{c|ccc|cc}
    \hline \thickhline
    \rowcolor{mygray}
    &\multicolumn{3}{c|}{REVERIE} &\multicolumn{2}{c}{R2R}  \\
    \cline{2-6}
    \rowcolor{mygray}
    \multicolumn{1}{c|}{\multirow{-2}{*}{$\lvert{{\Omega}_{\rm{n}}}\rvert$}}  &{SR}$\uparrow$ &SPL$\uparrow$ &{RGS}$\uparrow$ &{SR}$\uparrow$ &SPL$\uparrow$ \\
    \hline
    \hline
                               4    &53.30  &35.90  &\textbf{34.16}  &73.75 &62.77  \\
                               9    &\textbf{55.98}  &\textbf{39.66}  &33.71  &\textbf{75.80} &\textbf{65.37}  \\
                               16   &52.18  &36.37  &33.14  &73.82 &63.49  \\
                               25   &53.49  &35.87  &33.46  &74.63 &63.24  \\
    \hline
    \end{tabular}
    }
    \end{center}
        \vspace*{-15pt}
    \captionsetup{font=small}
	\caption{\small{Ablation study of neighborhood range on \textit{val} \textit{unseen} of REVERIE~\cite{qi2020reverie} and R2R~\cite{AndersonWTB0S0G18} (see \S\ref{ex_ablation} for more details).}}
    \label{table_neighborhood}
    \vspace*{1pt}
\end{table}

\begin{table}[t]
    \begin{center}
            \resizebox{0.49\textwidth}{!}{
            \setlength\tabcolsep{7pt}
            \renewcommand\arraystretch{1.05}
    \begin{tabular}{c|cc|cc|c}
    \hline \thickhline
    \rowcolor{mygray}
    &\multicolumn{2}{c|}{Occupancy} &\multicolumn{2}{c|}{Detection}&Layout  \\
    \cline{2-6}
    \rowcolor{mygray}
    \multicolumn{1}{c|}{\multirow{-2}{*}{Models}} &{IoU}$\uparrow$ &{mIoU}$\uparrow$ &mAP$\uparrow$ &mAR$\uparrow$ &{3D IoU}$\uparrow$ \\
    \hline
    \hline
    BEVFormer~\cite{li2022bevformer}    &20.38 &8.97 &27.30 &43.88 &62.71 \\
    OccNet~\cite{tong2023scene}        &22.12 &10.66 &29.91 &47.15 &64.04 \\
    \textbf{Ours}              &\textbf{24.31} &\textbf{12.93}  &\textbf{33.57}  &\textbf{51.60} &\textbf{66.45} \\

    \hline
    \end{tabular}
    }
    \end{center}
        \vspace*{-15pt}
    \captionsetup{font=small}
	\caption{\small{Quantitative results on 3D occupancy, 3D detection, and room layout prediction (see \S\ref{ex_perception} for more details).}}
    \label{table_3dtask}
    \vspace*{-3pt}
\end{table}

\vspace{10pt}
\subsection{Analysis on 3D Representation Learning}\label{ex_perception}

\noindent\textbf{Evaluation Metric.} Following standard protocols, we employ Intersection over Union (IoU) to evaluate the occupancy prediction quality, regardless of the semantic labels. The mean IoU (mIoU) of $15$ classes is also used to assess the performance of semantic occupancy. For 3D object detection, we utilize mean Average Precision (mAP) and mean Average Recall (mAR) with IoU thresholds of $0.50$. For room layout, we adopt 3D IoU for cuboid layout evaluation.

\vspace{1pt}
\noindent\textbf{Performance on 3D Tasks.} In Table~\ref{table_3dtask}, our network (\S\ref{sec_verencode}) outperforms other methods~\cite{li2022bevformer,tong2023scene} by a significant margin ($\textbf{2.19}\%$ on IoU of occupancy, $\textbf{3.66}\%$ on mAP of 3D detection, and $\textbf{2.41}\%$ on 3D IoU of room layout estimation). The mIoU of occupancy is also exhibits improvement ($\textbf{2.27}\%$), underscoring the network's proficiency in capturing both scene geometry and fine-grained semantics.

\vspace{1pt}
\noindent\textbf{Coarse-to-Fine Extraction.} Table~\ref{table_coarse2fine} lists the scores with different up-sampling operations (Eq.~\ref{eq_upsample}). Our approach improves the performance by solid margins (\eg, $11.03\%\!\!\rightarrow\!\!\textbf{12.93}\%$ for 3D occupancy, $75.14\%\!\!\rightarrow\!\!\textbf{75.80}\%$ on SR of R2R). This verifies the efficacy of our design of the coarse-to-fine extraction and learnable up-sampling operations.

\begin{table}[t]
    \begin{center}
            \resizebox{0.495\textwidth}{!}{
            \setlength\tabcolsep{7pt}
            \renewcommand\arraystretch{1.05}
    \begin{tabular}{c|ccc|cc}
    \hline \thickhline
    \rowcolor{mygray}
    &\multicolumn{3}{c|}{3D Perception} &\multicolumn{2}{c}{R2R}  \\
    \cline{2-6}
    \rowcolor{mygray}
    \multicolumn{1}{c|}{\multirow{-2}{*}{Up-Sampling}} &{mIoU}$\uparrow$ &mAP$\uparrow$ &{3D IoU}$\uparrow$ &{SR}$\uparrow$ &SPL$\uparrow$ \\
    \hline
    \hline
                                  \textit{w/o} Coarse-to-Fine     &12.39 &32.95 &\textbf{66.57} &$-$ &$-$ \\
                                   Trilinear Interpolation       &11.03 &29.42 &63.45 &75.14 &64.30 \\
                                   3D Deconvolution              &\textbf{12.93}  &\textbf{33.57}  &66.45  &\textbf{75.80} &\textbf{65.37} \\

    \hline
    \end{tabular}
    }
    \end{center}
        \vspace*{-15pt}
    \captionsetup{font=small}
	\caption{\small{Ablation study of \textit{Coarse-to-Fine Extraction} on occupancy prediction (mIoU), 3D detection (mAP), room layout (3D IoU), and \textit{val} \textit{unseen} set of R2R~\cite{AndersonWTB0S0G18} (see \S\ref{ex_perception} for more details).}}
    \label{table_coarse2fine}
    \vspace*{2pt}
\end{table}

\begin{table}[t]
    \begin{center}
            \resizebox{0.49\textwidth}{!}{
            \setlength\tabcolsep{5pt}
            \renewcommand\arraystretch{1.05}
    \begin{tabular}{ccc|ccc|cc}
    \hline \thickhline
    \rowcolor{mygray}
    \multicolumn{3}{c|}{Multi-task Learning} & \multicolumn{3}{c|}{3D Perception} & \multicolumn{2}{c}{R2R} \\
    \cline{1-8}
    \rowcolor{mygray}
    Occ. &Obj. &Room. &{mIoU}$\uparrow$ &mAP$\uparrow$ &{3D IoU}$\uparrow$ &{SR}$\uparrow$ &SPL$\uparrow$\\
    \hline
    \hline
    \checkmark   &             &                &12.09 &$-$ &$-$    &74.90 &63.82  \\
    \checkmark   &\checkmark   &                &12.14 &32.64 &$-$  &75.21 &64.79 \\
                 &\checkmark   &\checkmark      &$-$   &33.11 &64.58 &74.03 &63.51  \\
    \checkmark   &             &\checkmark      &11.37 &$-$  &63.29  &74.97 &64.66  \\

    \hline
    \checkmark   &\checkmark   &\checkmark      &\textbf{12.93}  &\textbf{33.57}  &\textbf{66.45}  &\textbf{75.80} &\textbf{65.37}  \\
    \hline
    \end{tabular}
    }
    \end{center}
        \vspace*{-11pt}
    \captionsetup{font=small}
	\caption{\small{Ablation study of \textit{Multi-task Learning} on occupancy prediction (mIoU), 3D detection (mAP), room layout estimation (3D IoU), and \textit{val} \textit{unseen} set of R2R~\cite{AndersonWTB0S0G18} (see \S\ref{ex_perception} for more details).}}
    \label{table_jointlearning}
    \vspace*{-2pt}
\end{table}

\vspace{1pt}
\noindent\textbf{Multi-task Learning.} Table~\ref{table_jointlearning} reports performance comparison with different perception tasks (\S\ref{sec_verencode}). We find that multi-task learning yields a substantial performance gain. This suggests these 3D perception tasks are complementary to each other in capturing geometric and semantic properties of scenes, further facilitating the decision-making.

\vspace{10pt}
\section{Conclusion}
In this paper, we propose a Volumetric Environment Representation (VER), which aggregates the perspective features into structured 3D cells. Through coarse-to-fine feature extraction, we can efficiently perform several 3D perception tasks. Based on this comprehensive representation, we develop the volume state for local action prediction and the episodic memory for retrieving the global context. We demonstrate that our agent achieves promising performance on VLN benchmarks (R2R, REVERIE, and R4R).


\newpage
       \twocolumn[
        \centering
        \Large
        \textbf{\thetitle}\\
        \vspace{0.5em}\textit{Supplementary Material} \\
        \vspace{1.0em}
       ] 

{

	This document provides more details of our approach and additional experimental results, which are organized as follows:
    \begin{itemize}
    \setlength{\itemsep}{0pt}
        \item Additional details (\S\ref{sec_addresult})
        \item Model details (\S\ref{sec_modeldetail})
        \item Discussion (\S\ref{sec_discussion})
	\end{itemize}
}

\section{Additional details}\label{sec_addresult}
\noindent\textbf{List of Symbols.} The$_{\!}$ below$_{\!}$ table$_{\!}$~will be$_{\!}$ definitely$_{\!}$ added$_{\!}$ into$_{\!}$ our$_{\!}$ supplementary$_{\!}$ material.$_{\!}$ We$_{\!}$ will omit$_{\!}$ unnecessary$_{\!}$ subscripts$_{\!}$ for$_{\!}$ notational$_{\!}$ convenience.

\vspace{-10pt}
\begin{table}[H]
\centering
        \resizebox{0.46\textwidth}{!}{
		\setlength\tabcolsep{6pt}
		\renewcommand\arraystretch{1.0}
\begin{tabular}{c|c|c}
\hline \thickhline
\rowcolor{mygray}
Notation &Description &Index \\
\hline
\hline
$N$ &Number of candidates  &\S\ref{sec_action} \\
$\mathcal{X}$ &Volume state space  &\S\ref{sec_stateestimate} \\

$\mathcal{G}$ &Episodic memory graph  &\S\ref{sec_action} \\
$\mathcal{V}$ &Observed viewpoints  &\S\ref{sec_action} \\
$\mathcal{E}$ &Navigable connections  &\S\ref{sec_action} \\

$\mathcal{A}$ &Local action space  &\S\ref{sec_action} \\
$\mathcal{A}^{*}$ &Global action space  &\S\ref{sec_action} \\ \hline

$\bm{E}$ &Instruction embeddings   &\S\ref{sec_stateestimate}\&\ref{sec_action}; Eq.~(\ref{eq_planeatt})\&(\ref{eq_globalaction}) \\
$\bm{Q}$ &3D volume query  &\S\ref{sec_verencode}; Eq.~(\ref{eq_sample}) \\
$\bm{F}^{\text{2d}}$ &2D perspective feature  &\S\ref{sec_verencode}; Eq.~(\ref{eq_sample}) \\
$\bm{F}^{\text{3d}}$ &3D volumetric representation  &\S\ref{sec_verencode}; Eq.~(\ref{eq_sample})\&(\ref{eq_upsample}) \\
$\bm{F}^{g}$ &Height-aware group representation  &\S\ref{sec_stateestimate}; Eq.~(\ref{eq_planeatt}) \\
$\bm{F}^{p}$ &Neighboring pillar representation  &\S\ref{sec_action}; Eq.~(\ref{eq_pillar}) \\
$\bm{G}$  &Node embeddings of $\mathcal{G}$  &\S\ref{sec_action}; Eq.~(\ref{eq_globalaction}) \\

$\bm{p}^{\text{3d}}$ &Local state transition probabilities  &\S\ref{sec_action}; Eq.~(\ref{eq_stateestimate}) \\
$\bm{p}^{\text{2d}}$ &Local action probabilities  &\S\ref{sec_action}; Eq.~(\ref{eq_localaction}) \\
$\bm{p}^{g}$ &Global action probabilities  &\S\ref{sec_action}; Eq.~(\ref{eq_globalaction})\&(\ref{eq_fuseaction}) \\
\hline
\multicolumn{3}{l}{~~~~\dag~Subscript $t$ in the paper denotes the navigation step.} \\
\end{tabular}
}
	\vspace*{-10pt}
\end{table}

\begin{figure}[t]
	\vspace{-5pt}
\begin{center}
	\includegraphics[width=0.99\linewidth]{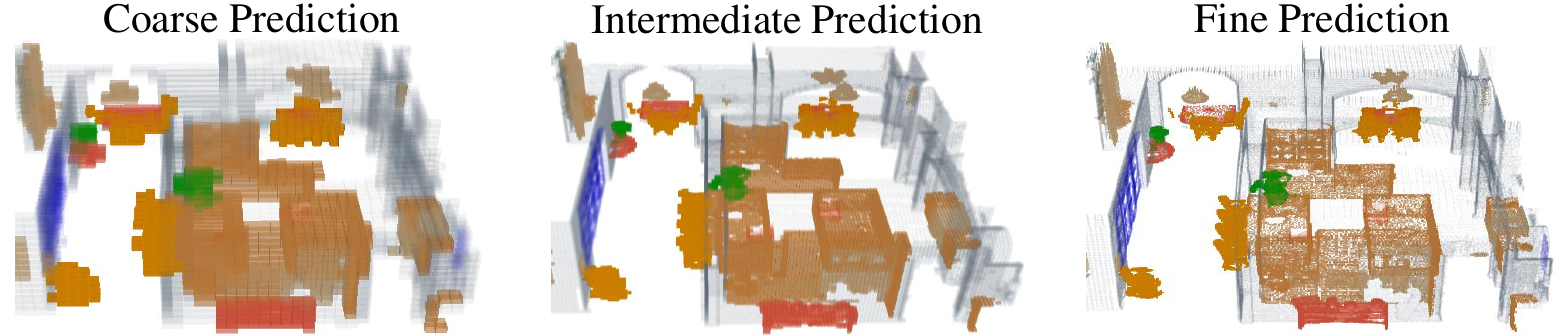}
\end{center}
\vspace{-20pt}
\captionsetup{font=small}
\caption{\small{Visualization of multi-resolution occupancy prediction (more details in \S\ref{sec_verencode}).}}
\label{fig_multires}
\vspace{-10pt}
\end{figure}

\noindent\textbf{Multi-resolution Labels.} In coarse-to-fine VER representation extraction (\S\ref{sec_verencode}), multi-resolution labels are utilized to supervise the perception network at each scale. The size of multi-resolution occupancy voxels are $0.4$m, $0.2$m, and $0.1$m, respectively. The layout estimation and object detection are also employed at each scale. Fig.~\ref{fig_multires} shows the coarse-to-fine occupancy prediction.

\begin{figure*}[t]
		
	\begin{center}
		\includegraphics[width=0.95\linewidth]{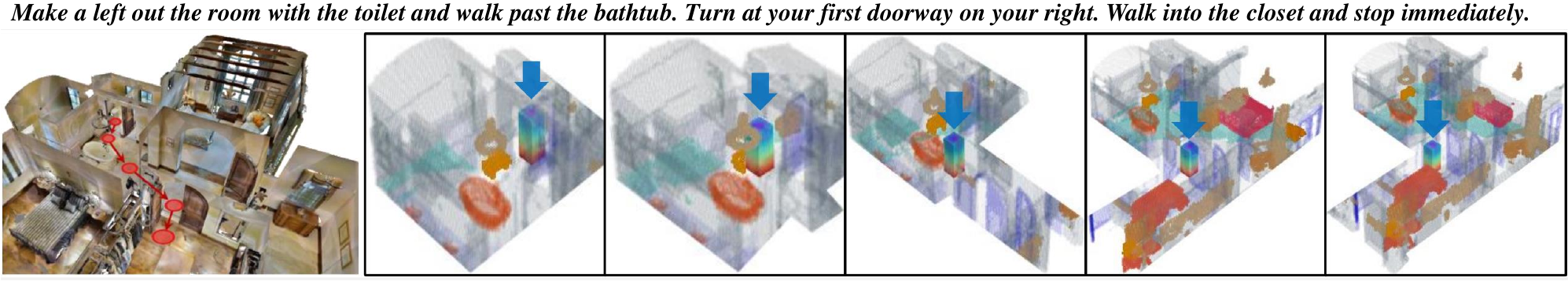}

	\end{center}
	\vspace{-20pt}
	\captionsetup{font=small}
	\caption{\small{Visual results on \textit{val unseen} of R2R (\textit{i},\textit{ii}) and REVERIE (\textit{iii}). During navigation, our agent recognizes the surrounding objects, captures the fine-grained details, and then performs comprehensive decision-making to finish the task successfully. Please zoom in for best view (more details in \S\ref{sec_addresult}).}}
	\label{fig_addvis}
	\vspace{-8pt}
\end{figure*}

\begin{figure*}[t]
		
	\begin{center}
		\includegraphics[width=0.95\linewidth]{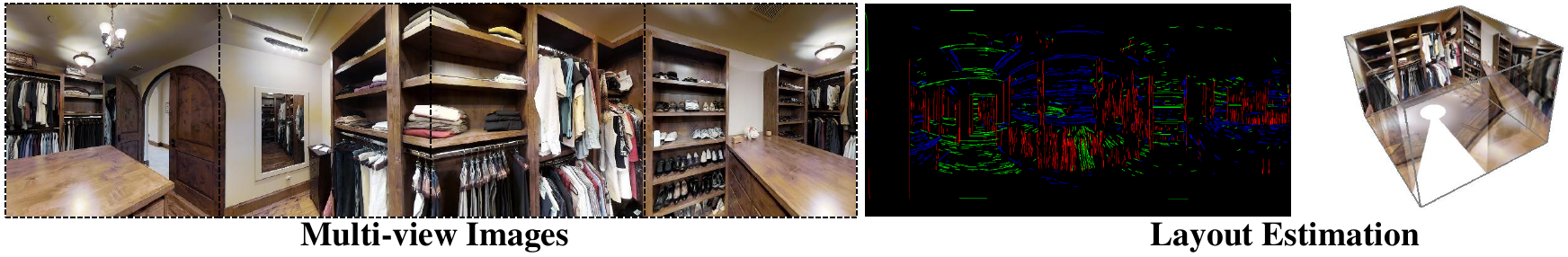}

	\end{center}
	\vspace{-20pt}
	\captionsetup{font=small}
	\caption{\small{Visualization of 3D room layout. Different from panoramic images in previous studies~\cite{zou2018layoutnet,zou2021manhattan}, we adopt multi-view 2D images as input. Please zoom in for best view (more details in \S\ref{sec_addresult}).}}
	\label{fig_layout}
	\vspace{-8pt}
\end{figure*}

\noindent\textbf{Visualization.} We provide more visualization results on \textit{val unseen} of R2R~\cite{AndersonWTB0S0G18} and REVERIE~\cite{qi2020reverie}. In Fig.~\ref{fig_addvis}, our agent recognizes the `toilet' and `bathtub', and then finds the first door easily. We illustrate the 3D layout estimation in Fig.~\ref{fig_layout}. Given the multi-view images as input, our model can capture the 3D geometric information and reconstruct the room structure.

\section{Model Details}\label{sec_modeldetail}
\subsection{Environment Encoder}
\noindent\textbf{Cross-view Attention(CVA).} We propose cross-view attention for 2D-3D sampling (\S\ref{sec_verencode}) and use the camera projection function $\mathcal{P}_{\text{c}}$ to obtain the reference points (Eq.~\ref{eq_sample}), which is formulated as follows:
\vspace{-2pt}
\begin{equation}\small
\begin{aligned}
{\text{CVA}}\big(\bm{Q}(x,y,z),{\bm{F}^{\text{2d}}}\big)=\text{DA}\big(\bm{Q}(x,y,z),\mathcal{P}_{\text{c}}(\bm{p}),\bm{F}^{\text{2d}}\big),
\end{aligned}
\vspace{-2pt}
\end{equation}
where $\bm{Q}(x,y,z)\in\mathbb{R}^{D_e}$ is located at $(x, y, z)$ position of $\bm{Q}\in\mathbb{R}^{D_e\times{X}\times{Y}\times{Z}}$, $\bm{F}^{\text{2d}}\in\mathbb{R}^{D_i\times{H}\times{W}}$ is the image feature, and $\mathcal{P}_{\text{c}}$ employs the camera intrinsic and extrinsic parameters for transformation. DA is the deformable attention~\cite{zhu2020deformable}:
\vspace{-2pt}
\begin{equation}\small
\begin{aligned}
{\text{DA}}\big(\bm{q},\bm{p},\bm{F}\big)=\sum_{k=1}^{K}\bm{W}_k\sum_{s=1}^{S}\bm{A}_{ks}\bm{W}_s\bm{F}(\bm{p}+\delta\bm{p}_{ks}),
\end{aligned}
\vspace{-2pt}
\end{equation}
where $K$ is the number of attention heads, $s$ indexes a total of $S$ sampling points, $\bm{W}_s$ is the learning weight, $\bm{A}_{ks}\in\left[0,1\right]$ is the learnable attention weight, $\delta\bm{p}_{ks}\!\in\!\!\mathbb{R}^2$ is the predicted offset to the reference point $\bm{p}$, and $\bm{F}(\bm{p}+\delta\bm{p}_{ks})$ is the feature at location $\bm{p}+\delta\bm{p}_{ks}$ computed by bilinear interpolation. We sample $S=6$ points for each query in CVA.

\noindent\textbf{Multi-task Learning.} We adopt ViT-B/16~\cite{dosovitskiy2020image} pretrained on ImageNet as the backbone. The size of the image features $\bm{F}^{\text{2d}}$ are $1280\times{1024}\times{768}$. We train the perception network with a detection head, a layout regression head, and a multi-class occupancy prediction head using the AdamW optimizer with a learning rate of $1\!\times\!10^{-4}$ for $500$ epoches (see \S\ref{sec_verencode}).

\subsection{Action Prediction}
\noindent\textbf{Object Prediction.} For REVERIE~\cite{qi2020reverie}, the agent needs to identify the specific objects. We first use the ViT-B/16 pretrained on ImageNet to extract the features of $N^o_t$ objects at $t$-th step $\bm{O}_t\!=\!\{\bm{o}_n|{\bm{o}_n}\!\in\!{\mathbb{R}}^{D_o}\}_{n=1}^{N^o_t}$, and add orientation features~\cite{chen2021history,chen2022think} with heading and elevation angles ($D_o=768$). Then these object features are concatenated with grouped VERs from different heights $\{\bm{F}^{g}_{t,z}\in\mathbb{R}^{{D_e}\times XY}\}^{Z}_{z=1}$ (\S\ref{sec_stateestimate}). We apply multi-layer transformers (MLT) to each group as (see Eq.~\ref{eq_planeatt}):
\begin{equation}
\small
\begin{aligned}
\widetilde{\bm{F}}^{g}_{t,z},\widetilde{\bm{O}}_{t,z}&=\text{MLT}\big([\bm{E};\bm{F}^{g}_{t,z};\bm{O}_{t,z}]\big),\\
{\widetilde{\bm{O}}_{t}}&=\sum\nolimits_{z=1}^{Z}{\widetilde{\bm{O}}_{t,z}}\in\!\mathbb{R}^{{D_o}\times {N^o_t}}.
\end{aligned}
\end{equation}

Then, the object prediction is formulated as:
\begin{equation}
\small
\begin{aligned}
\bm{p}^{o}_t&=\text{Softmax}(\text{MLP}({\widetilde{\bm{O}}_{t}}))\in\!\mathbb{R}^{N^o_t}.
\end{aligned}
\end{equation}

\noindent\textbf{Pretraining Objectives.} For R2R~\cite{AndersonWTB0S0G18} and R4R~\cite{jain2019stay}, we use Masked Language Modeling (MLM)~\cite{kenton2019bert,chen2021history} and Single-step Action Prediction (SAP)~\cite{hong2021vln,chen2021history} as auxiliary tasks in the pretraining stage. For REVERIE~\cite{qi2020reverie}, the Object Grounding (OG)~\cite{chen2022think,lin2021scene} is also used for object reasoning, and the sample ratio is MLM:SAP:OG=1:1:1. These auxiliary tasks are based on the input pair $(\bm{E},\bm{F}^{\text{3d}}_t,\bm{G}_t)$, where $\bm{E}\!\in\!\mathbb{R}^{D_w\times{L}}$ are the word embeddings, $\bm{F}^{\text{3d}}_t\!\in\!\mathbb{R}^{D_e\times{X}\times{Y}\times{Z}}$ is the encoded VER, and $\bm{G}_t\!\in\!\mathbb{R}^{D_e\times|{\mathcal{V}_t}|}$ are the node embeddings of the \textit{episodic memory} $\mathcal{G}_t$ at time step $t$ (see \S\ref{sec_approach}).

\noindent\textbf{Finetuning Objectives.} During finetuning, we alternatively use teacher-forcing and student-forcing for action prediction \cite{hong2021vln,chen2021history}. For REVERIE, OG is also adopted for finetuning:
\vspace{-1pt}
\begin{equation}
\small
\begin{aligned}
\mathcal{L}_{action}=0.25\mathcal{L}_{\rm {tf}}+\mathcal{L}_{\rm {sf}}+\mathcal{L}_{\rm {\scriptscriptstyle OG}}.
\end{aligned}
\vspace{-1pt}
\label{equ:finetune1}
\end{equation}


\section{Discussion}\label{sec_discussion}
\noindent\textbf{Terms of use, Privacy, and License.} Matterport3D~\cite{chang2017matterport3d}, R2R~\cite{AndersonWTB0S0G18}, and REVERIE~\cite{qi2020reverie} are available for non-commercial research purpose. Our code is implemented on the MMDetection3D codebase. MMDetection3D (\url{https://github.com/open-mmlab/mmdetection3d}) is released under Apache 2.0 license.

\noindent\textbf{Limitation.} As our agent is trained and evaluated on Matterport3D Simulator, where all environments are not dynamic, deploying the algorithm directly on a real-world robot may face challenges in capturing moving objects. Therefore, additional research and development are required to ensure safe deployment in real-world scenarios. This involves incorporating flow annotations and predicting voxel velocity for foreground objects. Our work primarily addresses the interior Vision-Language Navigation task. The generalization of this approach to other navigation tasks~\cite{krantz2020beyond,liu2023aerialvln,habitatchallenge2023} is not clear, and we plan to explore this in future work.

\noindent\textbf{Broader Impact.} We propose a powerful environment representation VER for VLN. Equipped with VER, our agent is able to perform comprehensive decision-making. On VLN benchmarks, our model demonstrates a promising improvement. In addition, we encourage more technical researching efforts devoted to environment representation learning for future research in the community.

{
    \small
    \bibliographystyle{ieeenat_fullname}
    \bibliography{main_arxiv}
}


\end{document}

%% file: preamble.tex
\usepackage[dvipsnames]{xcolor}
